%% file: paper.tex
\documentclass[conference]{IEEEtran}
\ifCLASSOPTIONcompsoc
  \usepackage[caption=false,font=normalsize,labelfont=sf,textfont=sf]{subfig}
\else
  \usepackage[caption=false,font=footnotesize]{subfig}
\fi
\usepackage{lipsum}
\usepackage{graphicx}
\usepackage[hidelinks]{hyperref}

\usepackage{todonotes}

\begin{document}

\input{tex/title}


\input{preamble/abstract}


\input{sections/introduction}

\input{sections/background}
\input{sections/method}
\input{sections/experiments}
\input{sections/results}
\input{sections/discussion}

\input{sections/future-work}
\input{sections/conclusion}




\input{./references/bibliography.tex}

\end{document}

%% file: tex/title.tex
\input{tex/docinfo}

\title{Deep Reservoir Computing \\Using Cellular Automata}

\author{\IEEEauthorblockN{Stefano Nichele}
\IEEEauthorblockA{Department of Computer Science\\
Oslo and Akershus University College of Applied Sciences\\
Oslo, Norway\\
stefano.nichele@hioa.no}
\and
\IEEEauthorblockN{Andreas Molund}
\IEEEauthorblockA{Department of Computer and Information Science\\
Norwegian University of Science and Technology\\
Trondheim, Norway\\
andrmolu@stud.ntnu.no}
}



\maketitle

%% file: tex/docinfo.tex
\newcommand{\reportTitle}{Deep Reservoir Computing using Cellular Automata}
\newcommand{\reportAuthor}{Andreas Molund}
\newcommand{\reportAuthorSupplement}{Department of Computer and Information Science\\
Norwegian University of Science and Technology\\
Trondheim, Norway\\
andrmolu@stud.ntnu.no}

%% file: preamble/abstract.tex

\begin{abstract}

Recurrent Neural Networks (RNNs) have been a prominent concept within artificial intelligence.
They are inspired by Biological Neural Networks (BNNs) and provide an intuitive and abstract representation of how BNNs work.
Derived from the more generic Artificial Neural Networks (ANNs), the recurrent ones are meant to be used for temporal tasks, such as speech recognition, because they are capable of memorizing historic input.
However, such networks are very time consuming to train as a result of their inherent nature.
Recently, Echo State Networks and Liquid State Machines have been proposed as possible RNN alternatives, under the name of Reservoir Computing (RC).
RCs are far more easy to train. 

In this paper, Cellular Automata are used as reservoir, and are tested on the 5-bit memory task (a well known benchmark within the RC community).
The work herein provides a method of mapping binary inputs from the task onto the automata, and a recurrent architecture for handling the sequential aspects of it.
Furthermore, a layered (deep) reservoir architecture is proposed.
Performances are compared towards earlier work, in addition to its single-layer version.

Results show that the single CA reservoir system yields similar results to state-of-the-art work. 
The system comprised of two layered reservoirs do show a noticeable improvement compared to a single CA reservoir. 
This indicates potential for further research and provides valuable insight on how to design CA reservoir systems.

\end{abstract}


%% file: sections/introduction.tex
 
\section{Introduction}


Temporal tasks, which we humans experience daily, are a great source of inspiration for research within the field of biologically-inspired artificial intelligence.
Systems capable of solving temporal tasks must be able to memorize historical data.
Recurrent Neural Networks (RNNs) are an example of a system of that sort, and have been studied for many years.
However, training RNNs is usually compute intensive.
One alternative is to consider recurrent networks as untrained reservoir of rich dynamics and only train an external feed-forward read-out layer.
The rich dynamics are to provide the necessary projection of the input features onto a discriminative and high dimensional space.
Basically, any substrate equipped with these properties can be used as a reservoir. This paper investigates the use of Cellular Automata (CA) computing substrates, inspired by \cite{yilmaz2014reservoir}.

CA at a microscopic scale are seemingly simple systems that exhibit simple physics, but at a macroscopic scale can reveal complex behavior which might provide the needed reservoir properties.
Specifically, CA are able to support transmission, storage, and modification of information \cite{langton1990computation}, all of which are necessary properties to support computation.

Furthermore, stacking reservoir systems in a multi-layered setup to offer additional computational capabilities have been successfully applied in \cite{triefenbach2010phoneme}, using a traditional RNN as reservoir.

The focus of the work herein is to explore series of CA reservoirs.
As such, a system with a single CA reservoir has been implemented first, and a second reservoir has been stacked at the end of the first one, to investigate whether two smaller layered reservoirs can replace a single greater one with regards to computational capacity.
The single CA reservoir system is therefore compared towards earlier work, as well as to the layered version.

The paper is organized as follows.
Section \ref{sec:background} presents background information.
Section \ref{sec:method} describes the specific method and system architecture in details.
Section \ref{sec:experiments} provides the experimental setup, and Section \ref{sec:results} outlines the experimental results.
A discussion is given in Section \ref{sec:discussion}.
Finally, Section \ref{sec:future} provides ideas for future work and Section \ref{sec:conclusion} concludes the paper.

%% file: sections/background.tex
\section{Background}
\label{sec:background}

\input{sections/background/reservoircomputing}
\input{sections/background/cellularautomata}
\input{sections/background/cainreservoir}


%% file: sections/background/reservoircomputing.tex

\subsection{Reservoir Computing}

\subsubsection{Fundamentals}


Information in feed-forward neural networks (NNs) is sent one way through layers of neurons; from an input layer, through one (or more) hidden layers, to an output layer.
Neurons in each layer are connected to neurons in the subsequent layer (except the last one) with weighted edges, and each neuron propagates signals according to its activation function.
A Recurrent Neural Network (RNN) contains the same basic elements.
However, it has recurrent connections that feed portions of the information back to the internal neurons in the network, making the RNN capable of memorization \cite{jaeger2002tutorial}, hence RNNs are promising architectures for processing of sequential tasks' data, e.g., speech recognition.
Ways of training RNNs are different variants of backpropagation \cite{werbos1990backpropagation,jaeger2002tutorial}, all with different computational complexity and time consumption.

One fairly recent discovery based upon the fundamentals of RNNs is Echo State Networks (ESNs) by Jaeger \cite{jaeger2001echo}. 
An ESN is a randomly generated RNN, in which the network does not exhibit any layer structure and its internal connection weights remain fixed (untrained), and can be treated as a reservoir of dynamics.
The "echo state property" is the activation state of the whole network being a function of previous activation states.
Training of such a network involves adapting only the weights of a set of output connections.

Another similar discovery is Liquid State Machines (LSMs) by Maas et al. \cite{maass2002real}.
It is similar to ESN in terms of topology, with an internal randomly generated neural network and problem-specific trained output weights. 
What differs is that LSMs are inspired by spiking networks that deal with noise or perturbations, whereas the ESNs are engineering-inspired and perform best without noise.

\begin{figure}[!t]
\centering
\includegraphics[width=1.3in]{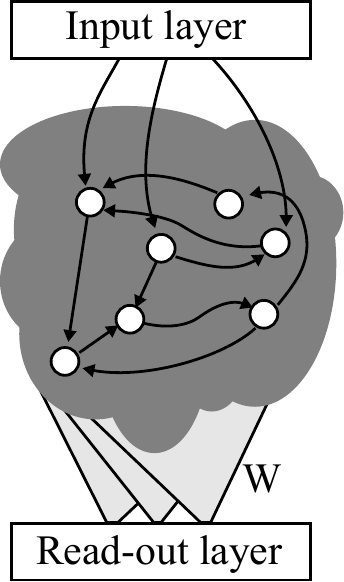}
\caption{A generic reservoir. There is only need to adapt weights $W$ towards a certain target.}
\label{fig:res}
\end{figure}

The basic idea of having readout nodes with trained weights connected to an arbitrary number of neurons inside the untrained reservoir, has been named \emph{reservoir computing} (RC).
\figurename\ \ref{fig:res} depicts this general RC idea.



\subsubsection{Physical Reservoir Implementations}

Different physical substrates have been shown to posses the necessary rich dynamics to act as reservoir.
Potentially, any high dimensional dynamic medium or system that has the desired dynamic properties can be used.
For example, in \cite{nikolic2006temporal}, a linear classifier was used to extract information from the primary visual cortex of an anesthesized cat.
In \cite{fernando2003pattern}, waves produced on the surface of water were used as an LSM to solve a speech recognition task.
The genetic regulatory network of the Escherichia Coli bacterium (E. coli) was used as an ESN in \cite{dai2004genetic} and as an LSM in \cite{jones2007ecoli}.
In \cite{dale2017reservoir,dale2016evolving,dale2016reservoir} unconventional carbon-nanotube materials have been configured as reservoir through artificial evolution. An optoelectronic reservoir implementation is presented in \cite{paquot2011optoelectronic,larger2012photonic}.


\subsubsection{Deep Reservoirs}

Within the RC research field, it has been suggested that reservoir performances may be improved by stacking multiple of them \cite{jalalvand2015real,triefenbach2010phoneme,jalalvand2015robust}. A critical analysis of deep reservoir systems is given in \cite{gallicchio2016deep}.
In a deep reservoir system, since the hidden units in the reservoir are not trainable, the reservoir's read-out values are sent as input to the next reservoir.
Thus, the reservoir and its associated output layer are stacked in a multi-layered (possibly deep) architecture.
This technique is inspired by Deep Neural Networks, in which adding layers of hidden units increases the ability of representation and abstraction, and thus the performances of the system.
One argument for stacking multiple reservoir systems is that the errors of one reservoir may be corrected by the following one, which may learn the semantics of the pattern that it gets as input.
As an example, in \cite{triefenbach2010phoneme} a deep reservoir architecture based on ESN proved successful in phoneme recognition.


%% file: sections/background/cellularautomata.tex

\subsection{Cellular Automata}

\begin{figure}[!t]
\centering
\subfloat[]{
  \includegraphics[width=1.5in]{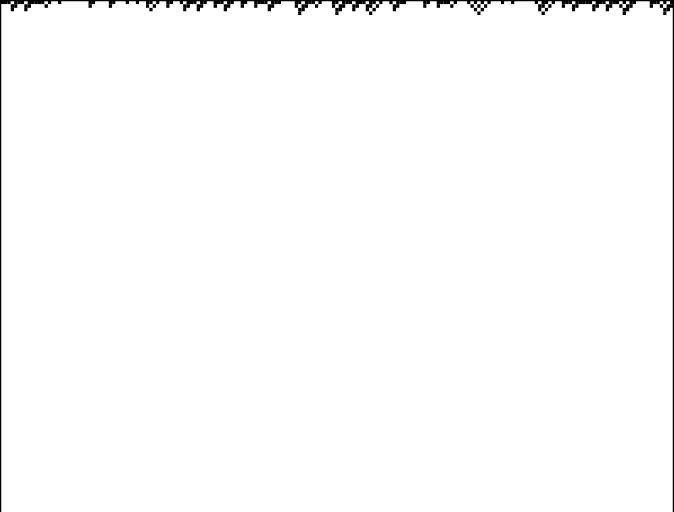}
  \label{fig:caclassi}}
~
\subfloat[]{
  \includegraphics[width=1.5in]{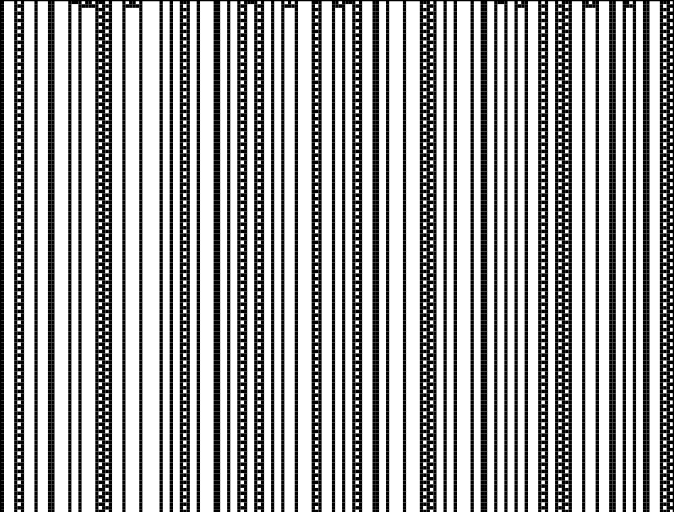}
  \label{fig:caclassii}}
\hfill
 \subfloat[]{
  \includegraphics[width=1.5in]{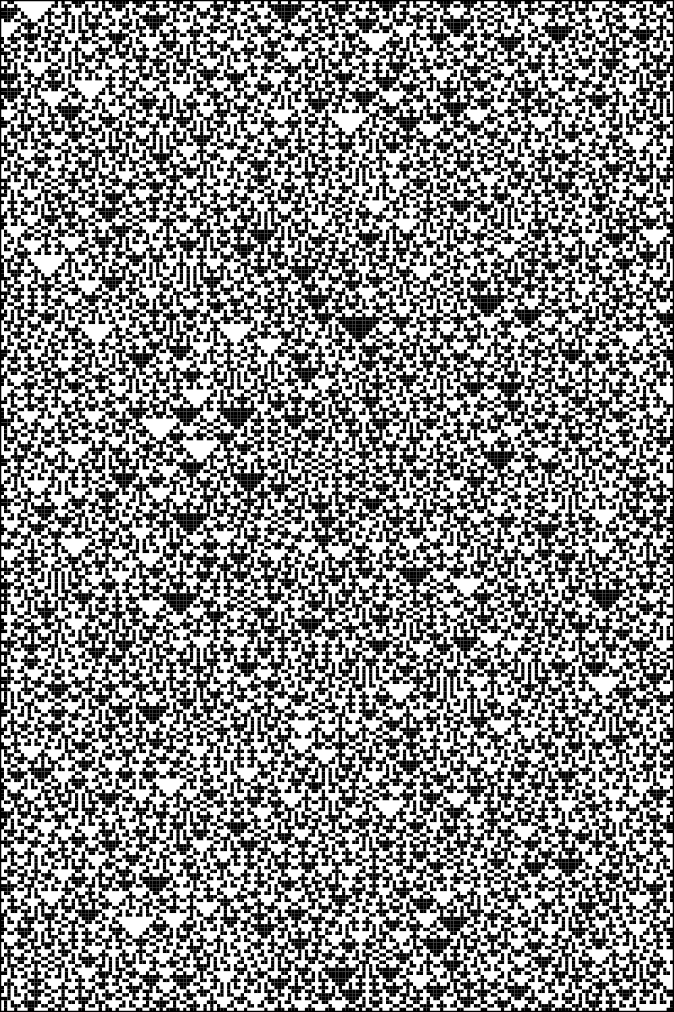}
  \label{fig:caclassiii}}
~
\subfloat[]{
  \includegraphics[width=1.5in]{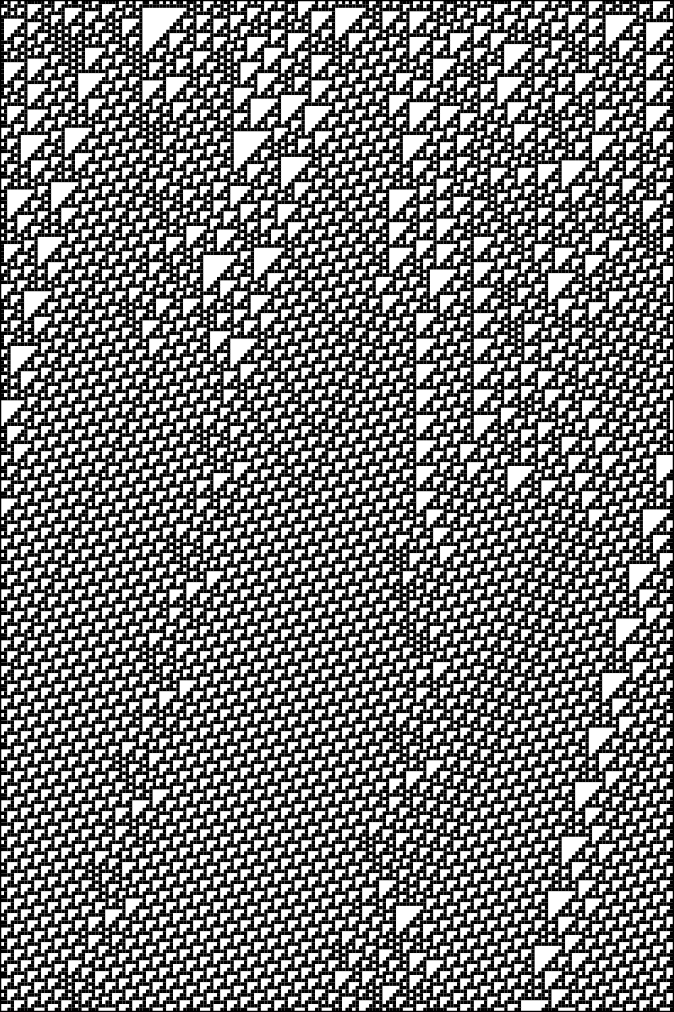}
  \label{fig:caclassiv}}
\caption{Elementary cellular automata iterating downwards. (a) and (b) are cut short. A black cell represents 1. These four are examples of each of Wolfram's classes: (a) is Class I with rule 40, (b) is Class II with rule 108, (c) is Class III with rule 150, and (d) is Class IV with rule 110.}
\label{fig:wolframclasses}
\end{figure}


Cellular Automata (CA) were inspired by the study of self-reproducing machines, by von Neumann in the 1940s \cite{von1966theory}.
CA are able to show emergent behaviour, i.e., the macroscopic properties are hard to explain from solely looking at the microscopic properties.
Within a cellular automaton, simple cells communicate locally over discrete time.
Locally means that a cell only interacts with its immediate neighbors, thus it has no global control.
The cells are discrete and placed on a regular grid of arbitrary dimension. The most common ones are 1D and 2D.
At each time step, all cells on the grid are updated synchronously based on their physics, i.e., a transition to a new state based on the previous state of the cell itself and its neighbors. Such transition tables are also referred to as CA rules.

Regarding the rule space, if $K$ is the number of states a cell can be in, and $N$ is the number of neighbors (including itself), then $K^N$ is the total number of possible neighborhood states.
Furthermore, each element is transitioning to one of $K$ states, thus, the transition function space is of size $K^{K^N}$.
For example, in a universe where cells have 5 possible states and three neighbors, there are $5^{5^3}\approx 2.4\times 10^{87}$ different rules or possible transition functions.

\begin{figure}[!t]
\centering
\includegraphics[width=\columnwidth]{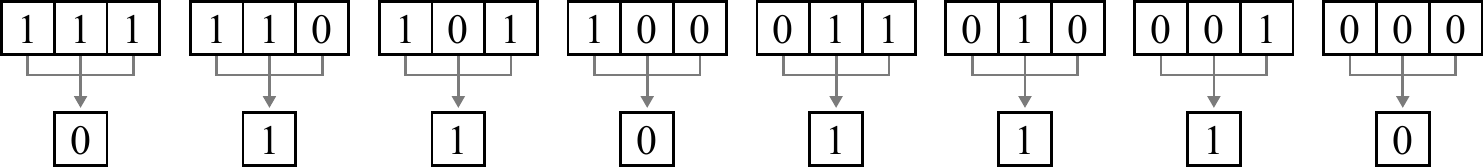}
\caption{The elementary CA rule 110. $(01101110)_2=(110)_{10}$}
\label{fig:elementaryca}
\end{figure}

Elementary CA are one of the simplest kind.
It comprises cells layed out in one dimension, in which $K=2$ and $N=3$.
The rule space can be enumerated in a base-2 system; each of the $2^{8}=256$ transition functions can be represented by a base-2 number of length $8$, as for example rule 110 in \figurename\ \ref{fig:elementaryca} that is represented as $(01101110)_2$.

Going a step in a more general direction, all one-dimensional CA were categorized by Wolfram \cite{wolfram1984universality} into four qualitative classes, based on the resulting evolution, i.e. the emergent CA behaviour.
Evolving 1D CA can easily be visualised by plotting the whole time-space diagram, iteration by iteration, downwards, see \figurename\ \ref{fig:wolframclasses} for illustrations.
CA in class I will always evolve to homogeneous cell states, independent of the initial states.
Class II leads to periodic patterns or single everlasting structures, either of which outcome is dependent on initial local regions of cell states.
Class III leads to a chaotic and seemingly random pattern.
Finally, class IV leads to complex localized structures which are difficult to predict, see \figurename\ \ref{fig:caclassiv}.

Langton introduced a scheme for parameterizing rule spaces in \cite{langton1990computation}, namely $\lambda$ parameter.
Briefly explained, within a transition function, the value of $\lambda$ represents the fraction of transitions that lead to a quiescent state.
As an example, rule 110 in \figurename\ \ref{fig:elementaryca} has $\lambda =0.625$.
If $\lambda =0.0$, then everything will transition to $0$, and the automaton will clearly lead to a homogeneous state.
$\lambda$ is especially useful for large rule spaces where it is hard to exhaustively enumerate all, because it can be used to generate rules with desired behavior.
Langton \cite{langton1990computation} did a qualitative survey throughout the rule space on 1D CA with $K=4$ and $N=5$;
rules were generated from different values of $\lambda$, from which CA were evolved and analyzed.
As the parameter increased from $0.0$ to $1.0$, the observed behavior underwent various phases, all the way from activity quickly dying out to fully chaotic.
In the vicinity of phase transition between ordered and chaotic, a subset of all CA rules was observed to lead to complex behavior that produced long-lasting structures and large correlation lengths.
Langton suggested that in this "edge of chaos" region is where computation may spontaneously emerge.



%% file: sections/background/cainreservoir.tex

\subsection{Cellular Automata in Reservoir computing}



CA reservoir have been first introduced in \cite{yilmaz2014reservoir}, and subsequently in \cite{yilmaz2015machine,bye2016investigation,margem2016how}. In \cite{gundersen2016cellular} the usage of non-uniform cellular automata has been proposed and in \cite{kleyko2017} a CA reservoir system has been used for modality classification of medical images.

Since the automata cells take on values from a discrete and finite set, mapping schemes to translate inputs onto CA may be needed.
For problems and tasks of binary nature such as 5-bit memory tasks \cite{hochreiter1997long} and temporal bit parity and density \cite{snyder2013computational}, this is relatively straightforward.
For input with real values, there are different proposed mapping methods \cite{yilmaz2014reservoir,kleyko2017,gundersen2016cellular}.
After translation, a rule is then applied to the automaton for some iterations, each of which is recorded so the nonlinear evolution becomes a projection of the input onto a discriminating state space.
This projection is later used in regression and classification for the task at hand.

CA as reservoirs provide several benefits over ESNs.
One is that the selection of reservoir, i.e., the CA transition table, is trivial; it is merely a choice of a CA rule with the wanted dynamics.
Even in elementary CA, one of the simplest form, there exists rules that are Turing complete, i.e. capable of universal computation \cite{cook2004universality}.
Another improvement is the aspect of computational complexity. 
According to \cite{yilmaz2014reservoir}, the speedups and energy savings for the N-bit task are almost two orders of magnitude because of the numbers and type (bitwise) of operations. Binarized variations of deep neural networks \cite{hubara2016binarized,umuroglu2016finn} and neural GPUs \cite{kaiser2015neural} have been recently suggested, in order to allow easier implementations in hardware than conventional deep neural network architectures. In \cite{fraser2017scaling} binarized neural networks are implemented on Field Programmable Gate Arrays (FPGAs). Such binary implementations are well suited for reconfigurable logic devices. One advantage of using binary CA (locally connected) over deep neural networks (fully connected) is a significantly lower memory cost and computation cost (binary operations implemented with a lookup table or bitwise logic in case of additive rules).


A vast sea of possibilities exists in how to set up a CA reservoir system.
For example, in a recent paper by Margem and Yilmaz \cite{margem2016how}, they explore memory enhancements of the CA by adopting pre-processing methods prior to evolution.
Further research with these possibilities can provide new understanding and insight in the field of Reservoir Computing with Cellular Automata (ReCA).


%% file: sections/method.tex

\section{Method}
\label{sec:method}

\begin{figure*}[!t]
\centering
\includegraphics[width=\textwidth]{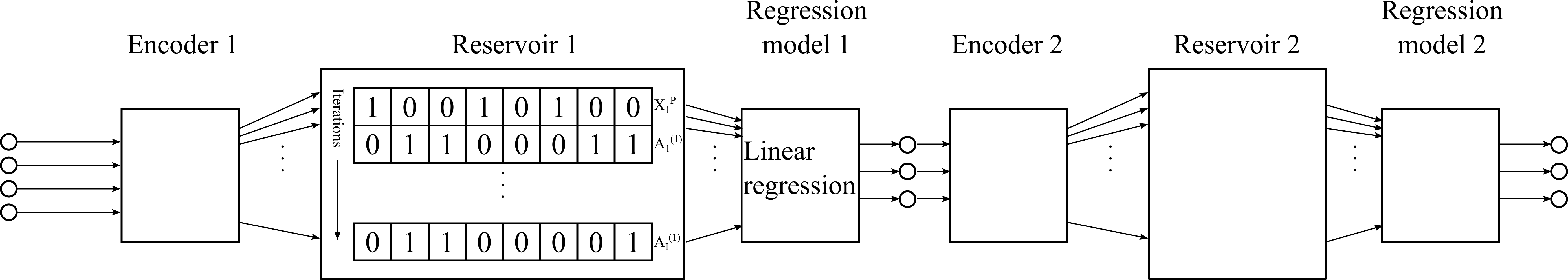}
\caption{System architecture.}
\label{fig:system-arch}
\end{figure*}


In this section, the used ReCA system is described in details.
The first implementation comprises a single reservoir tested (with several parameters) on the 5-bit memory task to compare to state-of-the-art results \cite{bye2016investigation,yilmaz2015machine}.
In the second implementation, an additional reservoir is added, which input is the output of the first one.
This larger system is tested on the same task (5-bit memory) for comparison.

\subsection{System Architecture}

Elementary cellular automata are used as the medium in the reservoirs, i.e., their cells have three neighbors (including itself), each of which can be in one of two states.
This means that there are 256 rules that can be applied, not all of which are used in this paper.
A selection of rules is presented in Section \ref{sec:results}.

\begin{figure}[!t]
\centering
\includegraphics[width=\columnwidth]{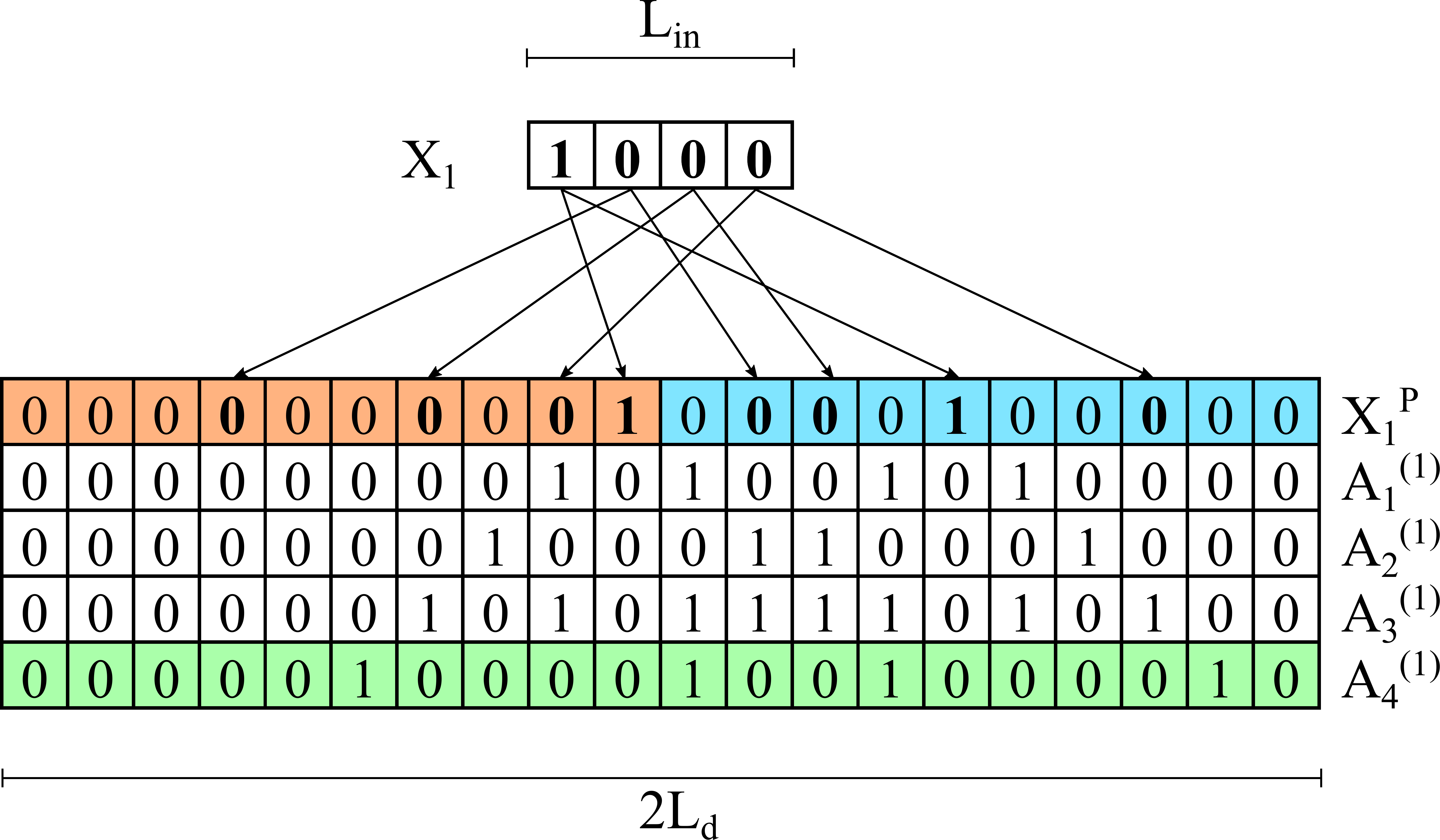}
\caption{Encoding input onto an automaton. $L_{in}=4$, $R=2$, $L_d=10$, $I=4$. The two different colors of $X_1^P$ signify the two different random mappings.}
\label{fig:encoding}
\end{figure}

In the encoding stage, input to the system is mapped onto the automata.  
Since the problem can be represented with binary vectors, the input elements translates directly into cell states.
This is one of the two input-to-automaton options proposed in \cite{yilmaz2014reservoir}, with the other one being for non-binary input data.
In addition to regular translation, the encoder is also responsible for padding, and diffusing the input onto an area that is of greater size than the input, if desired.
Padding is the method of adding elements of no information, in this case zeros, at some end of the mapped vector.
These buffers are meant to hold some activity outside of the area where the input is perturbing.
Thus, diffusing is a sort of padding by inserting zeros at random positions instead of at the end.
It disperses the input to a larger area.
The length of the area to which the input is diffused is denoted $L_d$.
Currently, out of these two methods of enlarging the memory capacity, only the diffuse parameter $L_d$ is set.
\figurename\ \ref{fig:encoding} illustrates how the system is mapping input to automata.

A reservoir can consist of $R$ separate cellular automata, each of which initial configuration is a randomly mapped input.
At system initialization, the indexes used for random mapping are generated, which means that the mappings are final and do not change throughout computation.
These automata are concatenated at the beginning of evolution to form a large initial configuration of size $R\times L_d$.
It is also possible to concatenate them after they are done iterating, but that proved to yield worse results.
At the boundaries, the automata are wrapped around, i.e., the rightmost cell has the leftmost cell as its right neighbor, and vice versa.

\begin{figure}[!t]
\centering
\includegraphics[width=\columnwidth]{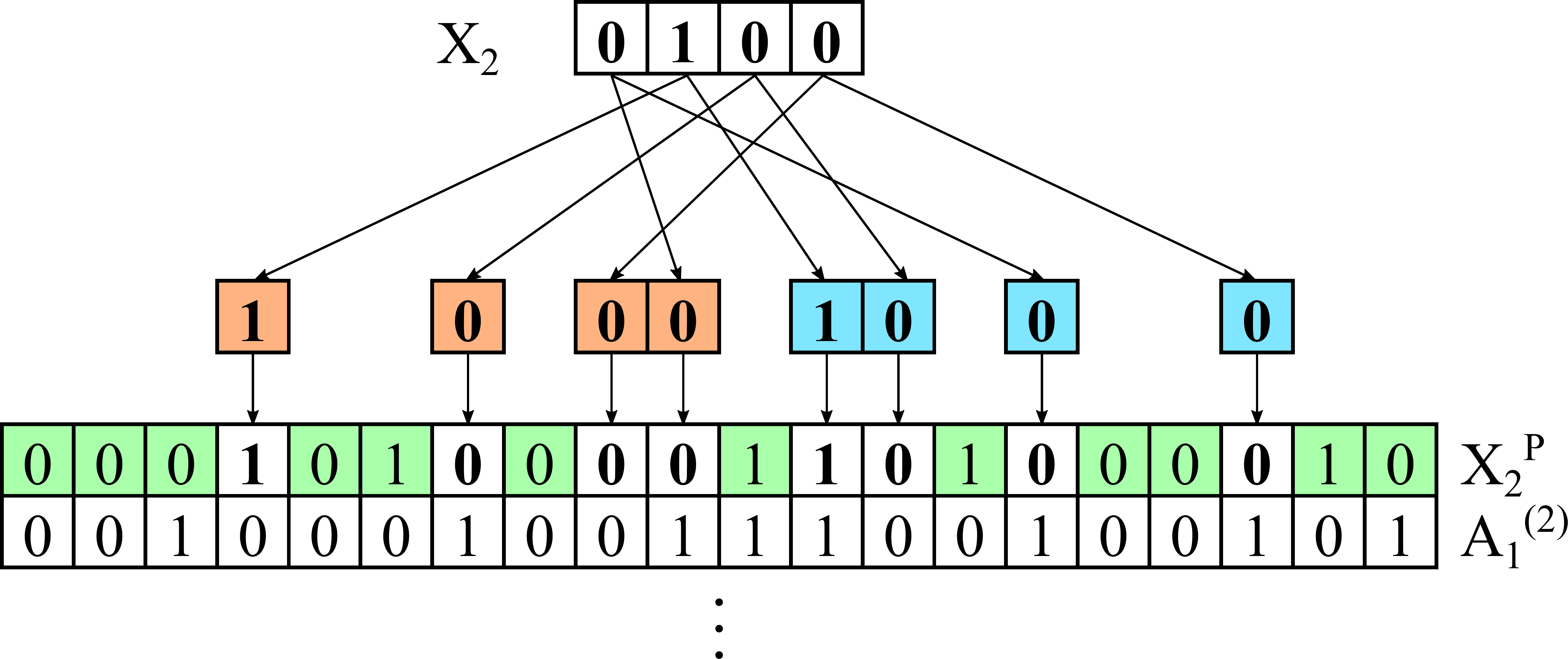}
\caption{Combining input with portions of the previous state. $X_2^P$ has traces of $A_4^{(1)}$ from \figurename\ \ref{fig:encoding}}
\label{fig:recurrent}
\end{figure}

For the 5-bit memory task described later in this paper, the system needs to be able to handle sequential inputs.
In \cite{yilmaz2015machine}, it is proposed a recurrent architecture for cellular automata in reservoir computing.

The system is initialized with an input $X_1$ at the first time step. 
$X_1$ is permuted, i.e. its elements are randomly mapped onto a vector of zeros according to the mapping scheme, $R$ times and concatenated to form the initial configuration of the automaton $X_1^P$:
$$X_1^P= [X_1^{P_1};X_1^{P_2};\ldots X_1^{P_R}]$$

The automaton is now of length $R\times L_d$. $Z$ is said to be the transition function, or rule, and is applied to the automaton for $I$ iterations. This renders an expressive and discriminative space-time volume of the input:
$$A_1^{(1)}=Z(X_1^P)$$
$$A_2^{(1)}=Z(A_1^{(1)})$$
$$\vdots$$
$$A_I^{(1)}=Z(A_{I-1}^{(1)})$$

$A_1^{(1)}$ through $A_I^{(1)}$ constitutes the evolution of the automaton, and is concatenated to form a state vector used for estimation at the first time step.
It is possible to include the permuted version of the input, i.e. the state before the first application of the rule, which for example is the case for the feedforward architecture in \cite{yilmaz2015machine}. 
However, it is excluded here:
$$A^{(1)}=[A_1^{(1)};A_2^{(1)};\ldots A_I^{(1)}]$$

Because this is a recurrent architecture, a fraction of the state vector is combined with the next input. 
Several methods exists; XOR, "normalized addition" which is adopted in \cite{yilmaz2015machine}, and a variant of overwriting which is implemented in \cite{bye2016investigation}.
For the subsequent time step, depicted in \figurename\ \ref{fig:recurrent}, the last iteration of the previous state vector is duplicated, after which the next input is permuted and written onto. 
In other words, instead of mapping the input onto a zero vector, it is done onto a vector that already contains information:
$$X_2^P=Y(X_2, A_I^{(1)})$$
where $Y$ is the function that overwrites $A_I^{(1)}$ with the permuted elements of $X_2$. 
One implication about this process is that the operation cannot be vectorized to an element-wise operand and hence hamper performance.
The positive is that no input information is lost, e.g. one of the input bits being zero (a signal is switched to off) will affect the subsequent evolution.
In other methods such as the probabilistic normalized addition, one relies on an increasing number of random mappings to increase the probability that input information is preserved.
To obtain the next state vector $A^{(2)}$, the transition function is applied on $X_2^P$ for $I$ iterations and concatenated:
$$A^{(2)}=[A_1^{(2)};A_2^{(2)};\ldots A_I^{(2)}]$$

$A^{(2)}$ is consequently used for estimation of time step 2. This process is repeated for every time step of the input sequence.


\subsection{Read-out}

As one can infer from what described earlier, the number of read-out values from the reservoir depends on the diffuse length, and the number of random mappings and iterations.
The read-out values from one time step is sent into a linear regression model together with its corresponding label.
Specifically the \texttt{linear\_model.LinearRegression} class from scikit-learn \cite{sklearn}.
For the ease of training, the model is fitted all at once with the output from all time steps for each element in the training set, together with their labels.
Even though the elements are from different time steps from different locations in the training set, they are weighted and treated equally because they each retain (to a greater or lesser degree) history from their respective "time lines".
Each corresponding label represent semantics from which the model is to interpret the read-out values. 

After the model is fit, it can be used to predict. 
Because linear regression is used, the output values from the predictions are floating points.
The output value $x$ is binarized according to Equation \ref{eq:binarize}.
\begin{equation}
x^b = \left\{ \,
\begin{IEEEeqnarraybox}[][c]{l?s}
\IEEEstrut
0 & if $x<0.5$, \\
1 & otherwise.
\IEEEstrut
\end{IEEEeqnarraybox}
\right.
\label{eq:binarize}
\end{equation}

\subsection{Deep CA Reservoir}

The reservoir computing framework described so far consists of one CA reservoir.
This approach is now expanded with new components, as depicted in \figurename\ \ref{fig:system-arch}, i.e., a second encoder, reservoir, and regression model.
After the values of the first read-out stage are classified, it is used as input to the second system.
Both regression models are fitted towards the same target labels.
One motivation for connecting two reservoir together is that the second can correct some of the mispredictions of the first one.

Training the system as a whole (and really testing it as well), involves the procedure that follows.
Inputs are encoded and mapped onto automata from which the first reservoir computes state vectors.
As this input is now transformed already, it is stored to be used for later prediction.
The first regression model is fitted with these feature vectors towards the corresponding labels, and immediately after, does his prediction.
These predictions are binarized, encoded, and mapped onto automata from which the second reservoir computes new state vectors.
Consequently, the second regression model follows the same training procedure as the first one, only with these new state vectors.
When the training is completed, the second regression model is tested for classification.


%% file: sections/experiments.tex

\section{Experiment}
\label{sec:experiments}


\subsection{5-bit Memory Task}

One sequential benchmark that has been applied to reservoir computing systems is the N bit memory task \cite{yilmaz2014reservoir,yilmaz2015machine,bye2016investigation,jaeger2007echo,gundersen2016cellular}, which is found to be hard for feedforward architectures \cite{hochreiter1997long}.

In the 5-bit memory task, a sequence of binary vectors of size four is presented to the system, where each vector represents one time step.
The four elements therein act as signals, thus, only one of them can be 1 at a time step.
This constraint also applies on the output which is also a binary vector, but rather with three elements.
In \cite{hochreiter1997long}, the problem was formulated with four output bits, but the fourth is "unused", hence it is omitted in the implementation herein.

For the first $5$ time steps in one run, the first two bits in the input vector are toggled between 0 and 1.
This is the information that the system is to remember.
If one of them is 1, the other one is 0, and vice versa, hence, there are a total of 32 possible combinations for the 5-bit task.
From $6$ throughout the rest of the sequences, the third bit is set to 1, except at time $T_d + 5$ where the fourth bit is set to 1.
The third bit is the distractor signal and indicates that the system is waiting for the cue signal (that is the fourth bit).
All runs presented in Section \ref{sec:results} are with $T_d=200$, meaning a total sequence length $T=T_d+2\times 5=210$.

\begin{figure}[!t]
\centering
\includegraphics[width=1.7in]{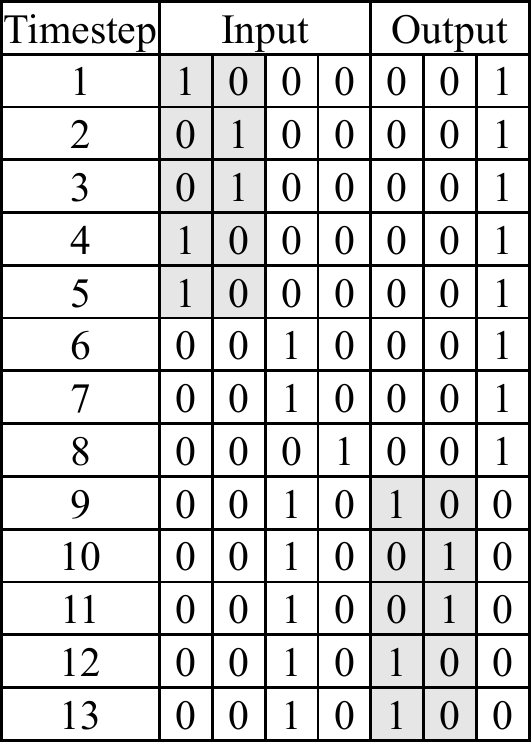}
\caption{An example of the 5-bit memory task with a distractor period $T_d=3$. The cue signal occurs at timestep 8, after which the first and second bit of the output are replicating the equivalent bits in the input (marked in gray).}
\label{fig:5-bit}
\end{figure}

As for the output, for all time steps until $T_d + 5$ inclusive, the third bit is 1.
Thereafter, the first and second bit are to replicate the $5$ first input signals.
See \figurename\ \ref{fig:5-bit} for an example.

\subsection{System Setup}

To collect the results, all 32 possible input patterns are used for both training and testing. 
Because the mappings are final throughout one run and hence the system is deterministic, the regression model is basically fitted towards the same output that it is to predict.
One run is said to be successful if the system can predict the right output for every time step for all of the 32 possible testing sets.
That means a total of $3\times 210\times 32$ bits correctly predicted.
Either 1000 or 100 runs is performed (and specified in the results figure), as explained in Section \ref{sec:results}.
The diffuse length is $L_d=40$.


%% file: sections/results.tex

\section{Results}
\label{sec:results}


The results of the 5-bit memory task are presented in this section and summarised in Table \ref{tab:res1} and Table \ref{tab:res2}.
Only the most promising rules have been tested with large combinations of $I$ and $R$.
The combination of $I$ and $R$ is denoted $(I,R)$ throughout the rest of the paper.

From (8,8) inclusive and up, the results are based on 100 runs.
Under that, they are based on 1000 runs, hence they are given with a single decimal precision.

Only a selection of all 256 possible rules is selected for experimentation.
Rules are selected based on available literature, e.g., \cite{bye2016investigation,yilmaz2014reservoir}, for comparison.
In \cite{yilmaz2014reservoir}, it is found that rules 22, 30, 126, 150, 182, 110, 54, 62, 90, and 60 are able to give 0 error for some combinations of $(I,R)$, where the best performing rules are 90, 150, 182, and 22, in decreasing order.
In \cite{bye2016investigation}, results are provided for rules 60, 90, 102, 105, 150, 153, 165, 180, and 195.

\begin{table*}[!t]
\caption{The correctness (\%) from the first reservoir computing system. Up until $(I,R)=(4,8)$ inclusive, 1000 runs were executed, hence the single decimal precision. With greater $I$ and $R$, only 100 runs.}
\label{tab:res1}
\centering
\begin{tabular}{|l|l|l|l|l|l|l|l|l|}
\hline
Rule & (I,R)=(2,4) & (2,8) & (4,4) & (4,8) & (8,8) & (8,16) & (16,8) & (16,16) \\ \hline
90   & 18.5        & 45.9  & 29.2  & 66.1  & 100   & 100    & 98     & 100     \\ \hline
150  & 0.2         & 1.8   & 6.7   & 33.7  & 89    & 100    & 100    & 100     \\ \hline
182  & 0.0         & 0.0   & 0.0   & 0.1   & 0     & 100    & 99     & 100     \\ \hline
22   & 0.0         & 0.0   & 0.0   & 0.0   & 0     & 99     & 100    & 100     \\ \hline
60   & 4.5         & 22.7  & 28.2  & 71.2  & 99    & 100    & 100    &         \\ \hline
102  & 6.0         & 24.0  & 28.1  & 69.7  & 97    & 100    &        &         \\ \hline
105  & 0.3         & 2.5   & 7.9   & 31.7  & 90    &        &        &         \\ \hline
153  & 3.1         & 20.2  & 28.9  & 70.6  & 99    &        &        &         \\ \hline
165  & 3.4         & 29.2  & 14.6  & 56.1  & 94    &        &        &         \\ \hline
180  & 0.0         & 0.0   & 0.0   & 0.0   & 0     &        &        &         \\ \hline
195  & 3.4         & 21.4  & 26.5  & 67.2  & 98    &        &        &         \\ \hline
\end{tabular}
\end{table*}

\begin{table*}[!t]
\caption{The correctness (\%) from the second reservoir computing system. What mentioned in the caption of Table \ref{tab:res1} still applies here.}
\label{tab:res2}
\centering
\begin{tabular}{|l|l|l|l|l|l|l|l|l|}
\hline
Rule & (I,R)=(2,4) & (2,8) & (4,4) & (4,8) & (8,8) & (8,16) & (16,8) & (16,16) \\ \hline
90   & 16.6        & 49.4  & 38.0  & 73.9  & 100   & 100    & 99     & 100     \\ \hline
150  & 0.3         & 3.5   & 10.4  & 39.7  & 90    & 100    & 100    & 100     \\ \hline
182  & 0.0         & 0.0   & 0.0   & 6.0   & 2     & 100    & 100    & 100     \\ \hline
22   & 0.0         & 0.0   & 0.0   & 0.0   & 0     & 100    & 100    & 100     \\ \hline
60   & 9.4         & 30.0  & 33.7  & 74.4  & 99    & 100    & 100    &         \\ \hline
102  & 9.8         & 31.9  & 35.2  & 71.9  & 97    & 100    &        &         \\ \hline
105  & 0.7         & 3.7   & 11.5  & 37.2  & 91    &        &        &         \\ \hline
153  & 5.0         & 24.6  & 35.4  & 73.9  & 99    &        &        &         \\ \hline
165  & 4.8         & 35.0  & 22.4  & 63.7  & 95    &        &        &         \\ \hline
180  & 0.1         & 0.2   & 0.1   & 0.1   & 0     &        &        &         \\ \hline
195  & 5.4         & 27.3  & 33.6  & 71.7  & 99    &        &        &         \\ \hline
\end{tabular}
\end{table*}

%% file: sections/discussion.tex

\section{Discussion}
\label{sec:discussion}

\subsection{Comparison with Earlier Work}

The individual results in \cite{bye2016investigation} are not quite equal to the equivalent in Table \ref{tab:res1}.
Some values differ noticeably, e.g., rule 90 and 165 at (4,4) which in \cite{bye2016investigation} result in a lower correctness.
The differences may be due to the different concatenation of random mappings, i.e., prior to CA evolution herein, whereas it is concatenated after in \cite{bye2016investigation}.
Furthermore, no rule is able achieve 100 \% correctness under (8,8), which is also the case in \cite{bye2016investigation}.

For the 5-bit task with a distractor period of 200, the best performing rule in \cite{yilmaz2015machine} needs a minimum of $(I,R)=(32,40)$ to produce 100 \% correct results.
That means a state vector of size (or trainable parameters) $32\times 40\times 4=5120$.
The method proposed herein needs $I\times R\times L_d=8\times 8\times 40=2560$, according to Table \ref{tab:res1}.


Some rules presented in Table \ref{tab:res1} are essentially equivalent.
Rule 102 is black-white equivalent with 153, i.e. they interchange the black and white cells, and left-right equivalent with rule 60, i.e. they interchange left and right cells.
The rule is furthermore both black-white and left-right equivalent with rule 195.
With these four rules being somehow equivalent, the obtained results in Table \ref{tab:res1} are also approximately equal.

It is furthermore experimented with padding versus diffusing, although not documented in this paper.
In the few executed tests, padding alone is observed to produce more stable results.
On the contrary, with only diffusion, the results seemed to yield better results overall but with higher variance.
The reason is most likely due to the larger area to which the input is mapped; individual elements of mapped input can both be very far apart and very close, while they are immediately adjacent when padding.

Another side experiment was executed to compare the fitting time when doubling the reservoir size.
The number of random mapping was doubled from 4 (i.e. number of trainable parameters or read-out nodes was $I\times R\times L_d=4\times 4\times 40=640$).
When doubling $R$ and hence doubling the state vector size, the outcome was an increase in fitting time by a factor of 3.4.
A set of 32 was used, each with 211 sequences.
It is a rough figure, but it gives an indication of the computational complexity.

\subsection{Layered Reservoirs}

It is quite intriguing how information represented by three bits from the first reservoir computing system can be corrected by the second one.
From a human's perspective, three bits is not much to interpret.
Nevertheless, the dynamics of the architecture and expressiveness of CA proves to be able to improve the result from a single reservoir system.

Table \ref{tab:res2} provides results on the correctness of the two-layer reservoir.
The results in each cell are directly comparable to the equivalent in Table \ref{tab:res1}, e.g. rule 150 at $(4,4)$ improved from 6.7 \% correctness to 10.4 \%.

Comparing the two tables, no rule in the second reservoir managed to get 100 \% before the first reservoir.
The rule that first is able to predict 100 \% correct is 90 at $(8,8)$, and in multi-layered setup, the same still applies; rule 90 at $(8,8)$.
Below this $(I,R)$ and where the first reservoir gets $>$ 10 \%, the best improvement gain is 53.4 \% for rule 165 at $(4,4)$.
Rule 165 also has the highest average improvement with its 21.7 \%.

Overall performance seems to increase with two reservoirs, the only decrease being rule 90 at the lowest $(I,R)$.
One possible explanation for the decrease, is that the reservoir has reached an attractor.
If it has, it must have occurred within the distractor period where the input signal does not change.
The system has been observed to reach an attractor when the addition method in the recurrent architecture is XOR, in which case the system reached an attractor after around two time steps within the distractor period.
However, the described implementation in this paper does not use XOR.

An intuitive comparison would be to compare whether adding a second reservoir of equal capacity can perform better than a single reservoir with twice the random mappings.
In that case, it does not seem to exist a viable configuration of rule and $(I,R)$.
However, when adding training time in the equation, a trade-off might be reasonable.

\begin{figure}[!t]
\centering
\subfloat[]{
  \includegraphics[width=3.2in]{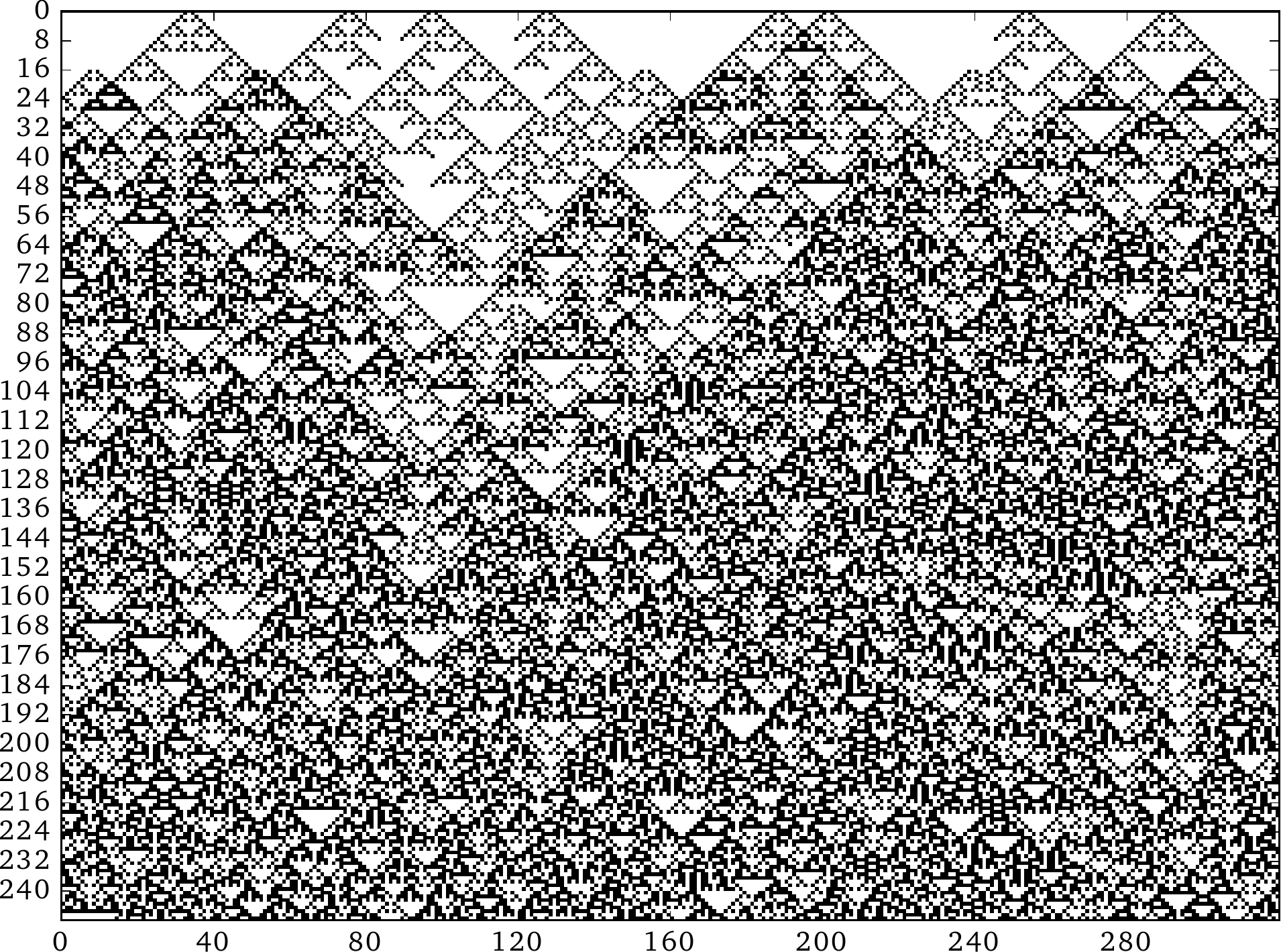}
  \label{fig:compute-example-res1}}
\vfill
\subfloat[]{
  \includegraphics[width=3.2in]{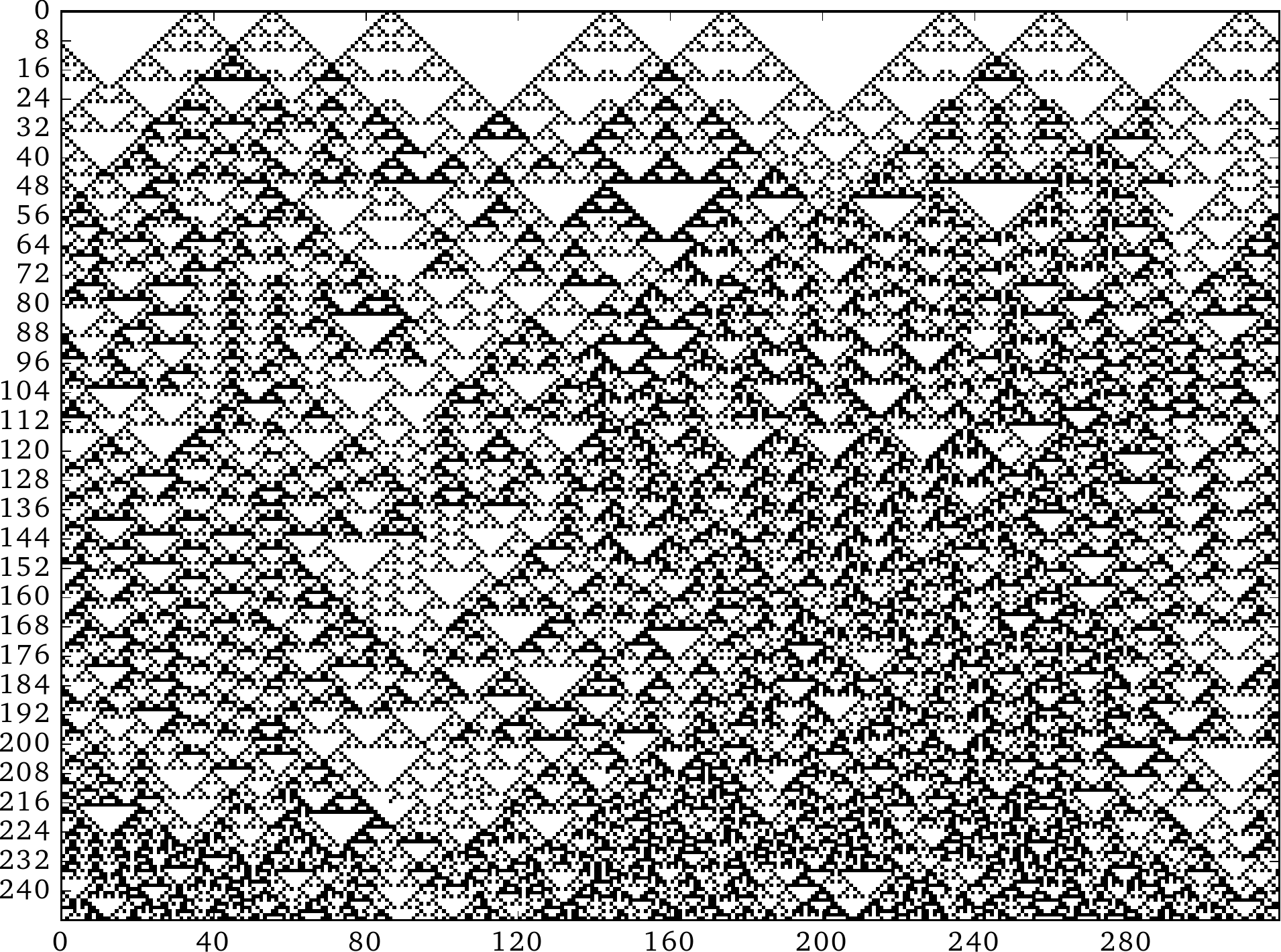}
  \label{fig:compute-example-res2}}
\caption{An example run on the 5-bit task with two reservoirs. (a) is the first reservoir and (b) is the second. $I=8$, $R=8$, $L_d=40$, and the distractor period is shortened to 20.}
\label{fig:compute-example}
\end{figure}

\figurename\ \ref{fig:compute-example} is a visualization of the actual CA states on a successful run on the 5-bit task with two reservoirs, although the distractor period is shortened down to 20 time steps.
Each tick on the vertical axis signifies the beginning of a new time step, and right before each tick, new input is added onto the CA state.
The input itself cannot be spotted, but the effects of it can (to a certain degree).
Spotting the input signals is feasible at early time steps, but gets more difficult at later iterations.

%% file: sections/future-work.tex

\section{Future Work}
\label{sec:future}






The results presented in this paper show that a system with two layered reservoirs performs better than a single reservoir.
This paper briefly touched upon one part of the vast spectrum of options and methods to opt for a practical implementation of ReCA systems. 
These options include the mapping method, investigating a larger rule space, the target of training for each regression model, the parameters for each reservoir, the rule of each reservoir, and the possibility of using 2-dimensional CA (e.g., Conway's Game of Life).
Especially interesting is the mapping method, or more generally, the preprocessing before exciting the medium within the reservoir.
For example, in \cite{margem2016how}, buffering and methods for handling subsequent inputs yield promising results.

One of the most interesting avenues for future work is to experiment further with more than two reservoirs, i.e., a deep ReCA system.

In addition, because of the nature of CA, ReCA systems are suitable for implementation in FPGAs. 
CA completely avoids floating point multiplications as opposed to ESNs, and furthermore, the regression stage can be replaced by summation \cite{yilmaz2014reservoir}.


%% file: sections/conclusion.tex

\section{Conclusion}
\label{sec:conclusion}

In this paper, a reservoir computing system with cellular automata serving as the reservoir was implemented. Such system was tested on the 5-bit memory task.
The system was also expanded to a two-layer reservoir, in which the output of the first reservoir inputs to the second reservoir.
Output of the first reservoir was used to compare the result with state-of-the-art work, as well as to the implementation with a layered reservoir. One of the main motivation for opting for a two-layered system is that the second reservoir can correct some of the mispredictions of the first one.

The results for the layered system show noticeable improvements when compared to the single reservoir system.
The greatest improvement (53.42 \%) was achieved by rule 165 at $(4,4)$.
Rule 165 proved to be promising in general, with an average improvement of 21.71 \%.


Overall, the second reservoir do improve the results of the first one to a certain degree.
Such improvement suggests that deep reservoir systems have the potential to become a viable solution.

%% file: references/bibliography.tex

\bibliographystyle{IEEEtran}
\bibliography{IEEEabrv,references/references}
